\documentclass{article}

\usepackage{PRIMEarxiv}

\usepackage[utf8]{inputenc} 
\usepackage[T1]{fontenc}    
\usepackage{hyperref}       
\usepackage{url}            
\usepackage{booktabs}       
\usepackage{amsfonts}       
\usepackage{nicefrac}       
\usepackage{microtype}      
\usepackage{lipsum}
\usepackage{fancyhdr}       
\usepackage{graphicx}       
\graphicspath{{media/}}     

\usepackage{algorithm}
\usepackage{algorithmic}
\usepackage{microtype}
\usepackage{graphicx}
\usepackage{subcaption}
\usepackage{booktabs} 
\usepackage{multirow}
\usepackage{enumitem}

\usepackage{amsmath}

\usepackage{amsmath}
\usepackage{amssymb}
\usepackage{mathtools}
\usepackage{amsthm}
\usepackage{xcolor}
\usepackage{arydshln}

\usepackage[capitalize,noabbrev]{cleveref}
\theoremstyle{plain}

\theoremstyle{definition}

\theoremstyle{remark}

\usepackage[textsize=tiny]{todonotes}

\title{Learning-State-Aware Dynamic Generative Data Augmentation on Small-Scale Datasets}

\author{
  Ting Xiang, Chenxi Deng, Jinhui Zhao, Bingting Jiang, Ke Zhang, Changjian Chen, Zhuo Tang \\
  Hunan University \\
  \texttt{\{txiang, changjianchen, ztang\}@hnu.edu.cn} \\
}

\begin{document}
\maketitle

\begin{abstract}

Small-scale image classification is often limited by the scarcity of training data. Generative data augmentation~(GDA) based on pretrained generative models has emerged as an effective solution. However, existing methods rely on task-agnostic augmentation strategies that overlook downstream model needs. Although recent dynamic GDA methods incorporate model feedback to guide augmentation, they still struggle to reliably determine sample-specific augmentation strengths and adapt augmentation strategies to different image regions while balancing image diversity and class semantics.

To address these issues, we propose learning-state-aware dynamic generative data augmentation~(LSADA). Specifically, LSADA constructs a learning state for each sample based on its current loss and loss-decrease rate, which is then mapped to a sample-specific augmentation strength. Furthermore, LSADA introduces a decoupled data augmentation and diffusion fusion strategy that applies strength-controlled transformations to class-relevant regions and generates diverse class-irrelevant regions, progressively fusing them to improve image diversity while preserving class semantics. Experiments on nine public datasets show that LSADA outperforms the existing SOTA dynamic GDA method by an average of 4.5\% on six natural image datasets and 2.5\% on three medical image datasets.

\end{abstract}


\section{Introduction}

The success of deep learning heavily relies on large-scale, high-quality datasets~\cite{deng2009imagenet}.
This reliance greatly limits the applicability of deep learning in small-scale data scenarios, where collecting and manually annotating sufficient training data is often costly and time-consuming~\cite{qi2020small}.
Recently, generative data augmentation (GDA) has emerged as a promising solution to this challenge by leveraging pretrained generative models (\textit{e.g.}, Stable Diffusion~\cite{rombach2022high}) to generate diverse training images. 
However, existing GDA methods often achieve limited performance gains because their generation process is independent of the downstream model. 
Consequently, the generated images may fail to address the specific weaknesses of the model being trained. 
For example, while a downstream classifier may struggle with ambiguous samples near its decision boundary or with underrepresented classes, a task-agnostic GDA method may continue to produce redundant and easily classified images.

\begin{figure}[t]
    \centering
    \includegraphics[width=.8\linewidth]{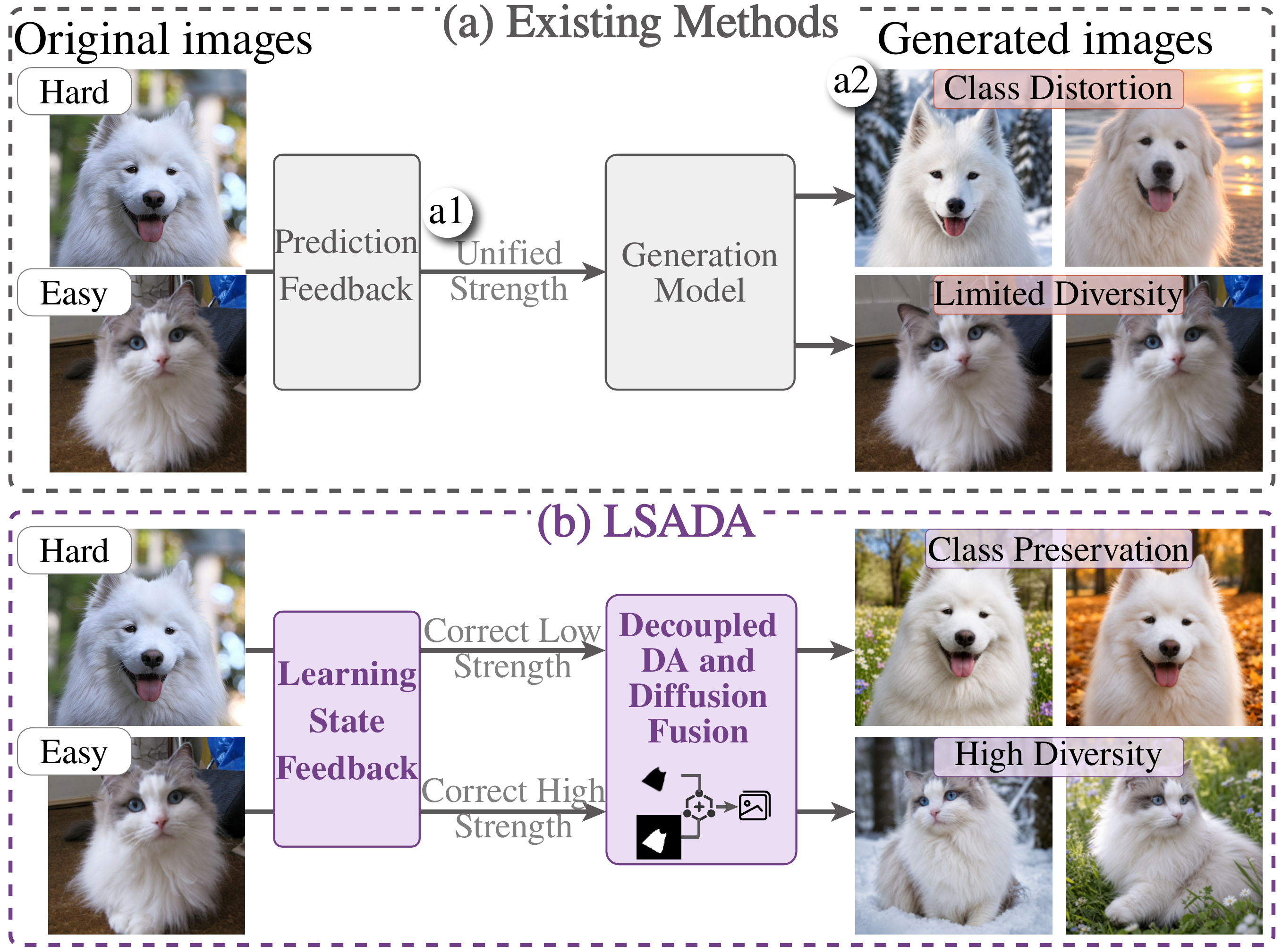}
    \caption{Comparison between (a)~existing dynamic GDA methods and (b)~our method, LSADA.}
    \label{fig:strength}
\end{figure}

To address this issue, dynamic GDA has been proposed, which uses downstream model prediction results~(\textit{e.g.}, uncertainty estimates or classification results) as feedback to select samples for further augmentation, such as ActGen~\cite{huang2024active} and DisCL~\cite{liang2025diffusion}.
Although effective to some extent, these dynamic GDA methods still face two key challenges.
\textbf{Challenge 1}~(Fig.~\ref{fig:strength}(a1)): different samples require different augmentation strengths. 
For example, easy samples may benefit from stronger augmentation to increase their difficulty, whereas hard samples typically require only mild augmentation to avoid underfitting.
However, the prediction-based feedback of existing GDA methods only determines which samples should be augmented with a unified strength.
Accurately estimating a sample-specific augmentation strength remains challenging, particularly in small-scale data scenarios, where model predictions during training are often unstable and unreliable.
\textbf{Challenge 2}~(Fig.~\ref{fig:strength}(a2)): even within a single image, different areas require different augmentation strengths.
For example, class-relevant regions should undergo mild augmentation to avoid class-semantic distortion, whereas class-irrelevant regions can be augmented more strongly to increase image diversity.
However, existing methods typically apply uniform augmentation across the entire image, overlooking such region-specific augmentation requirements.

To address these challenges, we propose \textbf{LSADA}, a learning-state-aware dynamic generative data augmentation method.
Instead of relying solely on prediction results, LSADA uses the current loss and loss-decrease rate to characterize each sample's learning state (\textbf{Challenge 1}).
The learning state is then mapped to a sample-specific augmentation strength, where larger values receive weaker augmentation to preserve semantic features, while smaller values receive stronger augmentation to improve diversity.
Then, LSADA introduces a decoupled data augmentation and diffusion fusion strategy to adapt to the derived augmentation strength (\textbf{Challenge 2}).
It applies strength-controlled rule-based transformations to class-relevant regions and uses LLM-guided prompts to generate diverse class-irrelevant regions.
The two components are progressively fused during reverse denoising, improving image diversity while preserving class semantics~(Fig.~\ref{fig:strength}(b)).
We experimentally validate the effectiveness of LSADA on nine public datasets.
Experiments show that our method outperforms the existing SOTA dynamic GDA method by an average of 4.5\% on six natural image datasets and 2.5\% on three medical image datasets.

In summary, the main contributions of our work are:

\begin{itemize}

    \item \textbf{A sample-specific augmentation policy} that dynamically adjusts augmentation strength according to the learning state of the downstream model.

    \item \textbf{A decoupled data augmentation and diffusion fusion generation strategy} that further ensures the diversity and class semantics of generated images.

    \item \textbf{A series of experimental results} demonstrates the effectiveness of our method.
\end{itemize}

\section{Related Work}

\begin{figure*}[!t]
    \centering
    \includegraphics[width=\textwidth]{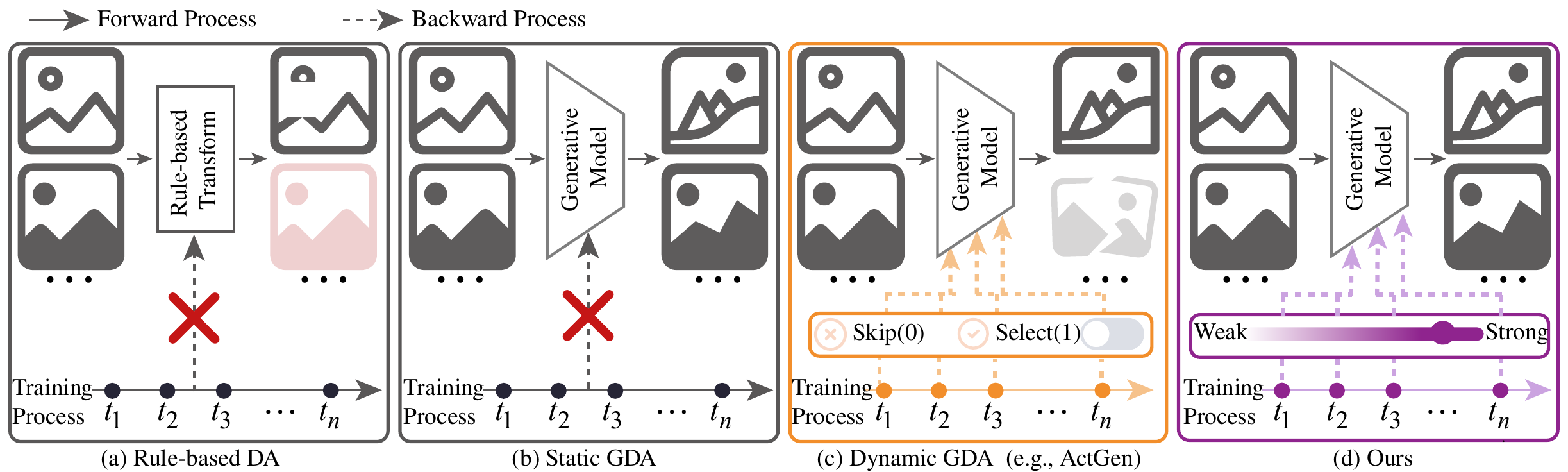}
    \caption{Evolution of data augmentation.}
    \label{fig:Related}
\end{figure*}

According to DA evolution~(Fig.~\ref{fig:Related}), existing methods can be broadly categorized into rule-based data augmentation and generative data augmentation.

\subsection{Rule-based Data Augmentation}

Rule-based data augmentation improves model generalization by applying predefined transformations to original images~(Fig.~\ref{fig:Related}(a)). 
Representative methods include CutOut~\cite{devries2017improved}, MixUp~\cite{hendrycks2019augmix,zhang2020does}, CutMix~\cite{yun2019cutmix}, AutoAugment~\cite{cubuk2018autoaugment}, and RandAugment~\cite{cubuk2020randaugment}.
Subsequent studies extended this paradigm by selecting sample-specific transformations, including OHL-Auto-Aug~\cite{lin2019online}, AdaAug~\cite{cheung2021adaaug}, InstaAug~\cite{miao2022learning}, MADAug~\cite{hou2023learn}, EntAugment~\cite{yang2024entaugment}, and AdaAugment~\cite{yang2025adaaugment}.
Although effective in certain scenarios, these methods are limited in generating diverse images because they only vary images at the pixel level.

\subsection{Generative Data Augmentation}

Generative data augmentation employs pretrained generative models, particularly diffusion models, to generate diverse images for downstream model training.
Depending on whether feedback from the downstream model is incorporated into the augmentation process, existing methods can be further categorized into static GDA and dynamic GDA.

\subsubsection{Static GDA.}
Static GDA~(Fig.~\ref{fig:Related}(b)) utilizes a generation model to generate samples before classifier training by perturbing latent features~\cite{antoniou2017data,islam2024diffusemix,
samuel2024generating,zhang2023expanding,zhu2024distribution}, optimizing prompts to guide generation~\cite{he2022synthetic,li2024semantic,
trabucco2024effective}, or editing image regions~\cite{islam2026instructmixup}.
Although effective, these methods are independent of downstream training and consequently may fail to generate samples that are fully beneficial to the downstream model.
To tackle this issue, dynamic generative data augmentation has been proposed recently.

\subsubsection{Dynamic GDA.}
Dynamic GDA~(Fig.~\ref{fig:Related}(c)) addresses this limitation by using feedback from the downstream model to control the augmentation process during training.
It adapts the generation process by optimizing the hyperparameters of an image-to-image~(I2I) generative model during training according to downstream model feedback.
The feedback includes classification results~\cite{koohpayegani2023genie, huang2024active}, sample uncertainty~\cite{jung2024dalda, wang2026difficulty, liang2025diffusion, askari2025improving}, sample utility~\cite{guo2026utilgen}, classification loss~\cite{nguyen2025we, li2024iterative}.

Although effective, these methods mainly use prediction-based feedback for sample selection and cannot determine sample-specific augmentation strengths. They also apply uniform augmentation, overlooking region-specific needs.
LSADA~(Fig.~\ref{fig:Related}(d)) instead estimates each sample's learning state from its current loss and loss-decrease rate and maps it to an augmentation strength.
It further introduces decoupled DA and diffusion fusion to separately augment class-relevant and class-irrelevant regions, improving diversity while preserving class semantics.

\section{Problem Formulation}

We first briefly introduce the dynamic GDA problem.
We are given a small labeled dataset $\mathcal{D}_o=\{(x_i,y_i)\}_{i=1}^{n}$ and a generative model $G$.
Each image $x_i\in\mathcal{X}\subseteq\mathbb{R}^{D}$ is associated with a label $y_i\in\mathcal{Y}=\{1,\ldots,c\}$, where $D$ is the input dimensionality and $c$ is the number of classes.

Unlike static GDA, dynamic GDA adapts image generation according to the feedback from the downstream classifier.
Let $f_{\theta^t}:\mathcal{X}\rightarrow\mathbb{R}^{c}$ denote the classifier at epoch $t$.
The generated data are updated every $k$ epochs.
Specifically, the generation epochs are defined as
$\mathcal{T}_g=\{k,2k,\ldots,Rk\}$,
where $R=\lfloor T/k\rfloor$ denotes the number of generation rounds and $T$ is the total augmentation training epochs.
At each generation epoch $t\in\mathcal{T}_g$, the feedback from $f_{\theta^t}$ determines the augmentation strength $a_i^t$ for each sample $x_i$.
The generative model $G$ then generates $m_t$ samples for each original sample:
\begin{equation}
    \mathcal{D}_{g_i}^{t}
    =
    \left\{
        \left(
            \hat{x}_{i,j}^{t}, y_i
        \right)
        \mid
        \hat{x}_{i,j}^{t}
        =
        G \left(
            x_i, y_i; a_{i}^{t}
        \right)
    \right\}_{j=1}^{m_t},
    \label{eq:dynamic_generated_subset}
\end{equation}
where $m_t$ is specified by the generation ratio.

The generated set at epoch $t$ is $\mathcal{D}_{g}^{t} = \bigcup_{i=1}^{n}\mathcal{D}_{g_i}^{t}$.
It is combined with the original data and the previously generated sets to update the classifier.
After all generation epochs, the complete generated dataset is $\mathcal{D}_{g}=\bigcup_{t\in\mathcal{T}_g}^{}{\mathcal{D}}_{g}^t$.
The overall generation ratio is $m=\sum_{t\in\mathcal{T}_g}m_t$, which indicates the total number of generated samples for each original sample.

Finally, the downstream classifier is trained by minimizing the empirical loss on both original and generated data:
\begin{equation}
    \theta
    =
    \arg\min_{\theta}
    \sum_{(x,y)\in\mathcal{D}_o\cup\mathcal{D}_g}
    \mathcal{L}\left(f_{\theta}(x),y\right),
\end{equation}
where $\mathcal{L}$ denotes the classification loss, such as cross-entropy.

\section{Method}

\begin{figure*}[!b]
    \centering
    \includegraphics[width=\textwidth]{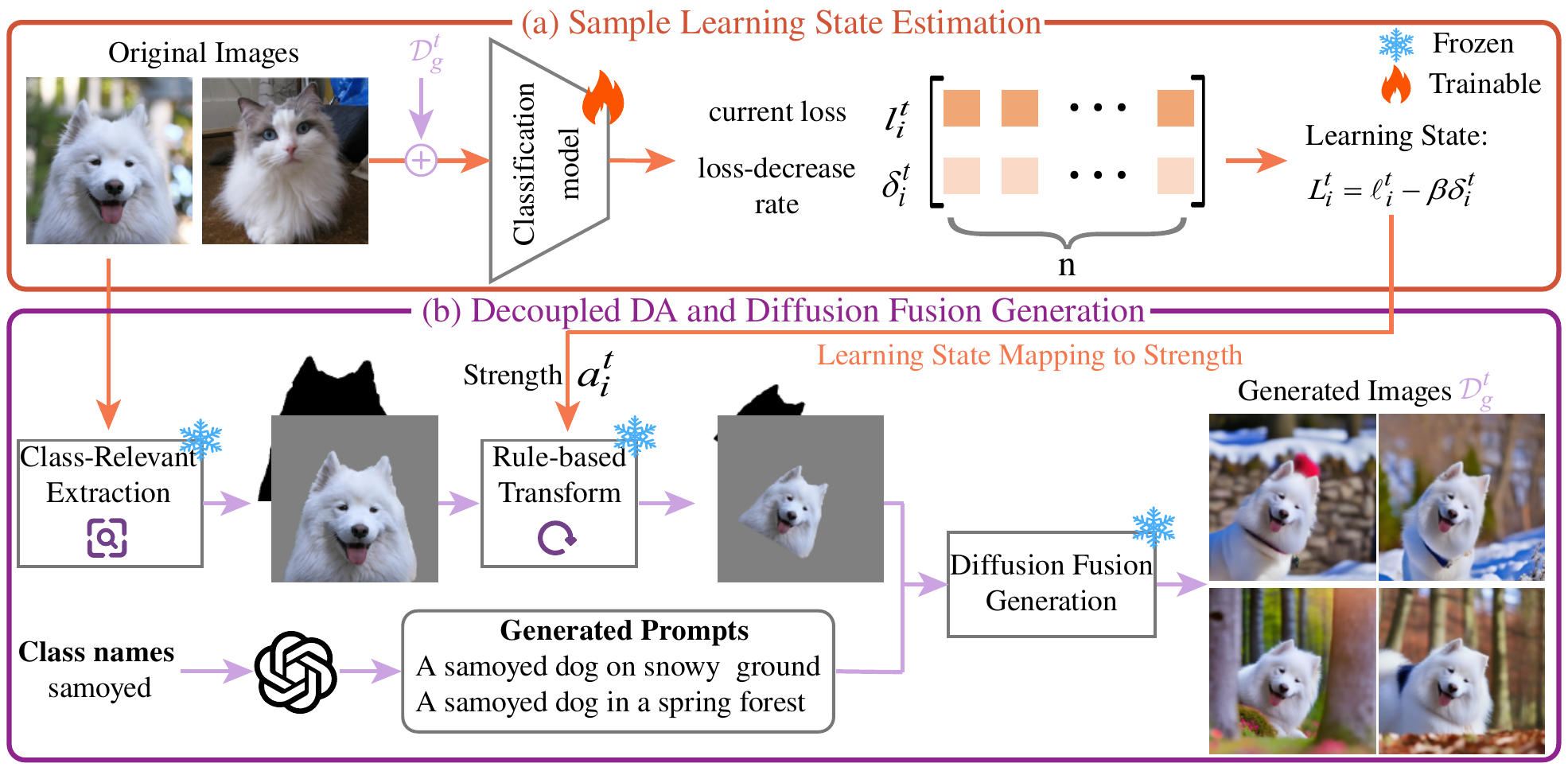}
    \caption{The framework of LSADA.
    LSADA consists of two stages: (a) sample learning state estimation, using current loss and loss-decrease rate, (b) decoupled data augmentation and diffusion fusion generation. The diffusion fusion generation pipeline is detailed in Fig.~\ref{fig:generation}. }
    \label{fig:framework}
\end{figure*}

We propose learning-state-aware dynamic generative data augmentation~(LSADA), which consists of two components: sample learning state estimation, and decoupled data augmentation and diffusion fusion generation.
First, LSADA estimates the learning state of each sample using its current loss and loss-decrease rate~(Fig.~\ref{fig:framework}(a)).
The estimated learning state is then mapped to a sample-specific augmentation strength, which controls rule-based transformations applied to the class-relevant region.
For the class-irrelevant region, we use a large language model to generate diverse prompts.
Finally, LSADA progressively fuses the transformed class-relevant region with the generated class-irrelevant region during the diffusion reversion process, generating diverse training images for the downstream classifier~(Fig.~\ref{fig:framework}(b)).
The overall algorithm is summarized in Appendix A.

\subsection{Sample Learning State Estimation}
\label{sec:learning_state}
Existing dynamic GDA methods typically use downstream model predictions as feedback to select original samples for further augmentation.
However, such feedback only determines which samples should be augmented, but cannot assign sample-specific augmentation strengths.
To construct a sample-specific augmentation strength, we characterize each sample's learning state using its current loss and loss-decrease rate, which reflect its current learning difficulty and recent learning progress, respectively~\cite{toneva2018an,paul2021deep}.

Specifically, we estimate the learning state of each sample every $k$ epochs.
At epoch $t$, we first compute the current loss of sample $(x_i,y_i)$ as
\begin{equation}
    \ell_i^t
    =
    \mathcal{L}\left(f_{\theta^t}(x_i), y_i\right),
    \label{eq:current_loss}
\end{equation}
where $f_{\theta^t}$ denotes the classifier at epoch $t$, and $\mathcal{L}$ denotes the classification loss, such as cross-entropy loss.

To measure how much each sample has been learned during the past $k$ epochs, we define the loss-decrease rate as
\begin{equation} 
\delta_i^t = {(\ell_i^{t-k}-\ell_i^t) }/{ (\ell_i^{t-k}+\epsilon_1) }, 
\label{eq:loss_decrease} 
\end{equation}
where $\epsilon_1$ is a small constant for numerical stability. 

Then we combine the current loss and the loss-decrease rate to define the learning state: 
\begin{equation} 
L_i^t = \ell_i^t - \beta \delta_i^t, 
\label{eq:learning_state} 
\end{equation} 
where $\beta>0$ balances the current learning difficulty and recent learning progress.
A larger $L_i^t$ indicates that the sample is harder to learn, characterized by a higher loss and slower improvement, while a smaller value suggests easier to learn.

\subsection{Learning-State-Aware Image Generation}
\label{sec:learning_state_aware_generation}
Given the estimated learning state of each sample, we further map it to an augmentation strength to guide image generation.

A straightforward solution is to use the augmentation strength to control the entire image generation process, such as adjusting the guidance scale to regulate the degree of image modification.
However, applying a unified strength to the entire image may cause class-semantic distortion or insufficient diversity.
To address this issue, we propose a decoupled data augmentation and diffusion fusion generation strategy, which decouples the augmentation of class-relevant and class-irrelevant regions, then fuses these components through diffusion fusion generation.

\subsubsection{Decoupled data augmentation.}
We first identify the class-relevant region in each image.
Given a class-relevant mask $M_i \in [0,1]^{H \times W}$, the class-relevant region of the original image $x_i$ is defined as
\begin{equation}
    x_i^{r} = M_i \odot x_i,
    \label{eq:relevant_content}
\end{equation}
where $H$ and $W$ denote the image height and width, respectively, and $\odot$ denotes element-wise multiplication.
For datasets with clearly defined class-relevant regions, we obtain $M_i$ using an off-the-shelf segmentation model.
For datasets without clearly defined class-relevant regions, we instead derive $M_i$ from class activation maps that highlight class-discriminative visual cues.

Then, we apply rule-based augmentations~(\textit{e.g.}, Rotation, Cutout, Resize) to the class-relevant region $x_i^{r}$, and control their strengths using the learning state $L_i^t$.
At epoch $t$, we first normalize the learning states $\{L_1^t, L_2^t, \ldots, L_n^t\}$ using normalization:
\begin{equation}
    \widetilde{L}_i^t
    =
    \frac{
        L_i^t - \min_j L_j^t
    }{
        \max_j L_j^t - \min_j L_j^t + \epsilon_2
    },
    \label{eq:normalized_learning_state}
\end{equation}
where $\epsilon_2$ is a small constant for numerical stability.

The normalized learning state is subsequently mapped to an augmentation strength:
\begin{equation} 
    a_i^t = a_{\min} + (a_{\max}-a_{\min})(1-\widetilde{L}_i^t), 
    \label{eq:strength_mapping} 
\end{equation} 
where $a_{\min}$ and $a_{\max}$ denote the minimum and maximum augmentation strengths, respectively.
Accordingly, a larger $\widetilde{L}_i^t$ leads to weaker augmentation, while a smaller $\widetilde{L}_i^t$ leads to stronger augmentation.

Based on the resulting strength, we apply the same rule-based transformation to both the class-relevant region $x_i^r$ and its mask $M_i$:
\begin{equation}
    \hat{x}_i^{r}
    =
    \mathrm{T}(x_i^{r}; a_i^t),
    \quad
    \hat{M}_i
    =
    \mathrm{T}(M_i; a_i^t),
    \label{eq:relevant_transform}
\end{equation}
where $\mathrm{T}(\cdot; a_i^t)$ denotes a rule-based transformation parameterized by the augmentation strength $a_i^t$.

To further enhance image diversity, rather than explicitly extracting class-irrelevant regions from the original image, we use a large language model~(LLM) to generate a set of diverse background prompts $\{p_i^j\}_{j=1}^{m_t}$, which guide the generation of class-irrelevant regions.
\begin{figure}[t]
    \centering
    \includegraphics[width=.8\linewidth]{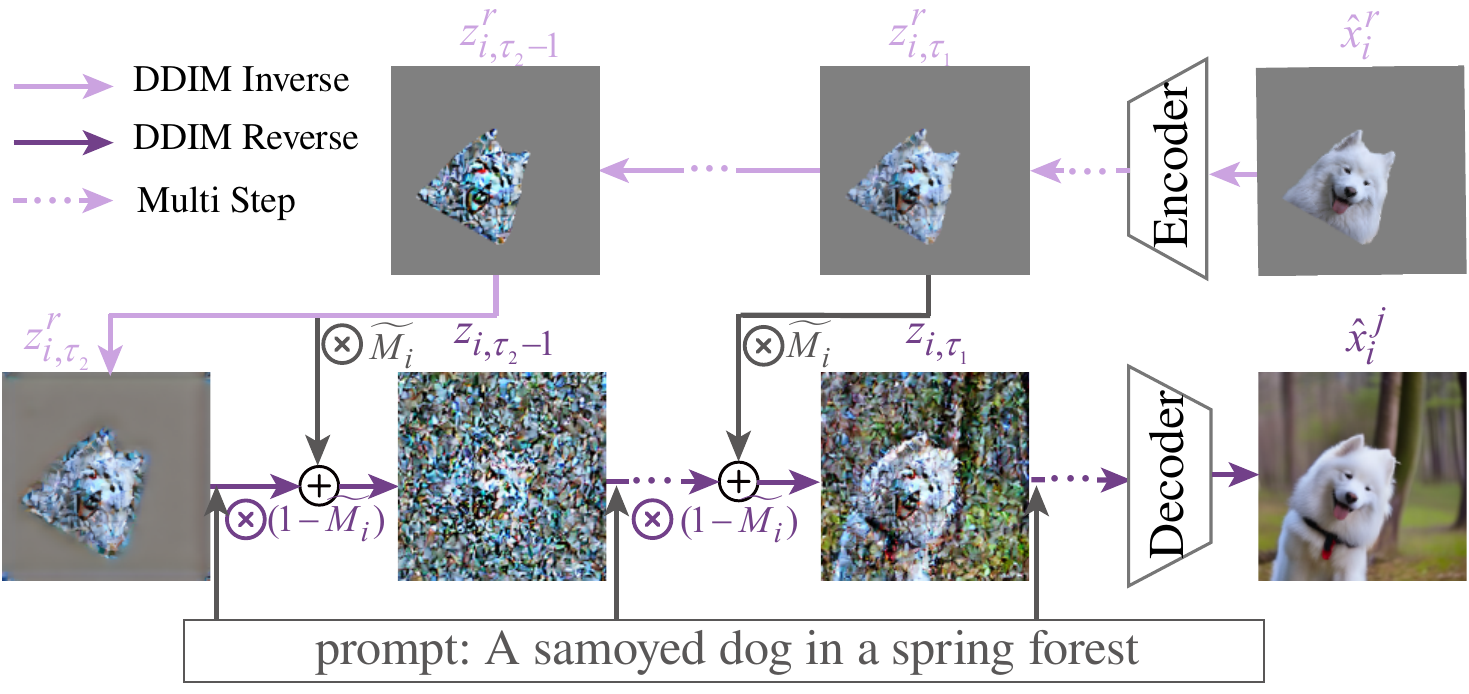}
    \caption{Diffusion fusion generation pipeline.}
    \label{fig:generation}
\end{figure}

\subsubsection{Diffusion fusion generation.} 
Given the augmented class-relevant region $\hat{x}_i^{r}$, the transformed mask $\hat{M}_i$, and a background prompt $p_i^j$, we fuse them to generate the image $\hat{x}_{i,j}^t$.

A straightforward way is to first generate a background image conditioned on the prompt using a Stable Diffusion model and then directly paste $\hat{x}_i^{r}$ onto it.
However, this approach may generate images that deviate from the real data distribution.
Moreover, the class-relevant region may not blend well with the generated background, which can degrade downstream classification performance.
To address these issues, we introduce a diffusion fusion generation process~(Fig.~\ref{fig:generation}).

To be specific, we first encode the augmented class-relevant region $\hat{x}_i^{r}$ into the latent space:
\begin{equation}
    z_{i,0}^{r} = \mathrm{Enc}(\hat{x}_i^{r}),
    \label{eq:latent_encode}
\end{equation}
where $\mathrm{Enc}(\cdot)$ denotes the VAE encoder.

Since diffusion fusion is performed in the latent space, we resize the transformed mask $\hat{M}_i$ to match the spatial dimensions of the latent feature map using nearest-neighbor interpolation:
\begin{equation}
\widetilde{M}_i
=
\mathrm{Resize}{H_z \times W_z}(\hat{M}_i),
\quad
\widetilde{M}_i \in [0,1]^{H_z \times W_z},
\label{eq}
\end{equation}
where $\mathrm{Resize}{H_z \times W_z}(\cdot)$ denotes nearest-neighbor resizing to the spatial resolution $H_z \times W_z$ of the latent feature map.

To preserve the semantic structure of the class-relevant region, we first apply DDIM inversion~\cite{song2020denoising} to the encoded latent $z_{i,0}^{r}$.
We record the inverted latent trajectory over the fusion interval $[\tau_1, \tau_2]$:
\begin{equation}
    \{z_{i,\tau}^{r}\}_{\tau=\tau_1}^{\tau_2}
    =
    \mathrm{Inv}(z_{i,0}^{r}),
    \label{eq:foreground_inversion}
\end{equation}
where $\mathrm{Inv}(\cdot)$ denotes the DDIM inversion process.

Then we initialize the reversion process with the inverted latent at timestep $\tau_2$:
\begin{equation}
    z_{i,\tau_2} = z_{i,\tau_2}^{r}.
    \label{eq:fusion_initialization}
\end{equation}
Starting from $z_{i,\tau_2}$, we perform reverse denoising conditioned on the background prompt $p_i^j$.
For each timestep $\tau \in [\tau_1,\tau_2]$, the latent is first updated by one reverse denoising step:
\begin{equation}
    z_{i,\tau-1}
    =
    \mathcal{R}_{\tau}(z_{i,\tau}, p_i^j),
    \label{eq:reverse_denoising_prompt}
\end{equation}
where $\mathcal{R}_{\tau}(\cdot, p_i^j)$ denotes the reverse denoising operation at timestep $\tau$ conditioned on prompt $p_i^j$.

After each denoising step, we inject the inverted class-relevant latent into the corresponding masked region:
\begin{equation}
    z_{i,\tau-1}
    =
    \widetilde{M}_i \odot z_{i,\tau-1}^{r}
    +
    (1-\widetilde{M}_i) \odot z_{i,\tau-1},
    \label{eq:masked_injection}
\end{equation}
where $\odot$ denotes element-wise multiplication.
The first term preserves the class semantics structure, while the second term allows the class-irrelevant region to be generated according to the background prompt.

We continue the reverse denoising process from $\tau_1$ to $0$ without masked injection:
\begin{equation}
    z_{i,\tau-1}
    =
    \mathcal{R}_{\tau}(z_{i,\tau}, p_i^j),
    \quad
    \tau \in [1,\tau_1] .
    \label{eq:standard_reverse_denoising}
\end{equation}

Finally, the latent at timestep $0$ is decoded to obtain the generated image:
\begin{equation}
    \hat{x}_{i,j}^{t}
    =
    \mathrm{Dec}(z_{i,0}),
    \label{eq:final_decode}
\end{equation}
where $\mathrm{Dec}(\cdot)$ denotes the VAE decoder.

\section{Experiments}

\subsection{Experimental Settings}

\subsubsection{Datasets.}
We evaluate the effectiveness and efficiency of LSADA on nine public small-scale image classification datasets, covering both natural and medical images.
The natural image datasets include Caltech-101~\cite{fei2004learning} and CIFAR100-Subset~\cite{krizhevsky2009learning} for coarse-grained classification; Cars~\cite{krause2013collecting}, Flowers~\cite{nilsback2008automated}, and Pets~\cite{parkhi2012cats} for fine-grained classification; and DTD~\cite{cimpoi2014describing} for texture classification.
The medical image datasets are drawn from MedMNIST~\cite{yang2021medmnist} and include PathMNIST~\cite{kather2019predicting} for colon pathology, BreastMNIST~\cite{al2020dataset} for breast ultrasound, and OrganSMNIST~\cite{xu2019efficient} for abdominal CT classification.
Detailed dataset statistics are provided in Appendix~B.

\subsubsection{Baselines.}
To evaluate the effectiveness of our proposed method, we compare LSADA with representative and state-of-the-art~(SOTA) data augmentation methods from different categories, 
including rule-based augmentation methods~(\textit{i.e.}, CutOut~\cite{devries2017improved}, RandAugment~\cite{cubuk2020randaugment}, TrivialAugment~\cite{muller2021trivialaugment}, TeachAugment~\cite{suzuki2022teachaugment}, MADAug~\cite{hou2023learn} and EntAugment~\cite{yang2024entaugment}), 
SOTA static generative augmentation methods~(GIF~\cite{zhang2023expanding}), 
and SOTA dynamic generative augmentation methods~(DisCL~\cite{liang2025diffusion} and ActGen~\cite{huang2024active}).

\begin{table*}[!b]
    \centering
    \resizebox{\linewidth}{!}
    {
    \begin{tabular}{lccccccccccc}
         \toprule
         \multirow{2}{*}{\textbf{Dataset}} & \multicolumn{7}{c}{\textbf{Natural image datasets}} & \multicolumn{4}{c}{\textbf{Medical image datasets}} \\
        \cmidrule(lr){2-8} \cmidrule(lr){9-12}
        & Caltech 101 & Cars & Flowers & DTD & CIFAR100-S & Pets & \textbf{Avg.} & PathMNIST & BreastMNIST & OrganSMNIST & \textbf{Avg.} \\
        \midrule
         original & 26.3 & 19.8 & 74.1 & 23.1 & 35.0 & 6.8 & 30.9 & 72.4 & 55.8 & 76.3 & 68.2\\

         CutOut & 51.5 & 25.8 & 77.8 & 24.2 & 44.3 & 38.7 & 43.7 & 78.8 & 66.7 & 78.3 & 74.6\\
         RandAugment & 57.8 & 43.2 & 83.8 & 28.7 & 46.7 & 48.0 & 51.4 & 79.2 & 68.7 & 79.6 & 75.8 \\
         TrivialAugment & 49.9 & 21.1 & 81.8 & 28.0 & 37.3 & 5.9 & 37.3 & 83.2 & 61.0 & 78.2 & 74.1 \\
         TeachAugment & 70.5 & 25.9 & 58.7 & 51.5 & 31.3 & 68.7 & 51.1 & 79.6 & 74.3 &  77.2 & 77.0 \\
         MADAug & 65.5 & 55.3 & 84.7 & 44.4 & 38.2 & 67.8 & 59.3 & 75.6 & 73.1 & 75.9 & 74.9 \\
         EntAugment & 70.7 & 69.0 & 75.9 & 40.8 & 52.4 & 67.8 & 62.8 & 86.2 & 74.4 & 77.7 & 79.4 \\
         \hdashline
         SD 2.1 & 55.7 & 64.5 & 80.7 & 32.4 & 55.9 & 42.4 & 55.3 & 76.3 & 73.7 & 51.6 & 67.2 \\
         GIF & 54.4 & 60.6 & 82.1 & 33.9 & 61.1 & 52.9 & 57.5 & 86.9 & 77.4 & 80.7 & 81.7 \\
         DisCL & 45.8 & 26.8 & 82.2 & 30.6 & 30.3 & 45.8 & 43.6 & 83.7 & 76.9 & 79.2 & 79.9 \\
         ActGen & 78.2 & 90.9 & 97.5 & 60.2 & 50.7 & 86.2 & 77.3 & 90.3 & 78.8 & 80.8 & 83.3 \\
         \shortstack[l]{Ours\\{\phantom{(+11.0)}}}
         & \shortstack{\textbf{86.1}\\{({+7.9})}}
         & \shortstack{\textbf{91.4}\\{({+0.5})}}
         & \shortstack{\textbf{98.3}\\{({+0.8})}}
         & \shortstack{\textbf{63.4}\\{({+3.2})}}
         & \shortstack{\textbf{61.7}\\{({+11.0})}}
         & \shortstack{\textbf{89.7}\\{({+3.5})}}
         & \shortstack{\textbf{81.8}\\{({+4.5})}}
         & \shortstack{\textbf{90.9}\\{({+0.6})}}
         & \shortstack{\textbf{84.6}\\{({+5.8})}}
         & \shortstack{\textbf{81.8}\\{({+1.0})}}
         & \shortstack{\textbf{85.8}\\{({+2.5})}} \\
         
         \bottomrule
    \end{tabular}
    }
    \caption{Accuracy (in \%) of a ResNet-50 trained from scratch on original and generated images by different methods. }
    \label{tab:overall_performance}
\end{table*}

\subsubsection{Implementation details.}
We use ResNet-50 as the downstream classifier for all datasets and train it for 200 epochs. Image generation is performed only during the first (T=100) training epochs, with the generated data updated every 5 epochs.
Following the GDA baselines, we use \textit{Stable Diffusion} v.2.1~(SD 2.1) for image generation and expand each dataset with a generation ratio $m=5$.
To extract class-relevant regions, we employ SAM3~\cite{carion2025sam} for Caltech-101, Cars, Flowers, CIFAR100-S, and Pets, and CAM~\cite{zhou2016learning} for the remaining datasets.
For class-relevant region transformation, we apply Rotation with angles in $[5^\circ,20^\circ]$, Cutout with ratios in $[5\%,20\%]$ and Resize with scaling ratios in $[60\%,100\%]$.
For class-irrelevant regions, we use the GPT-5.5 API to generate diverse background prompts.
More details are provided in Appendix~C.

\subsection{Overall Performance}

\subsubsection{DA effectiveness.}

\begin{figure*}[t]
  \centering
   \includegraphics[width=\textwidth]{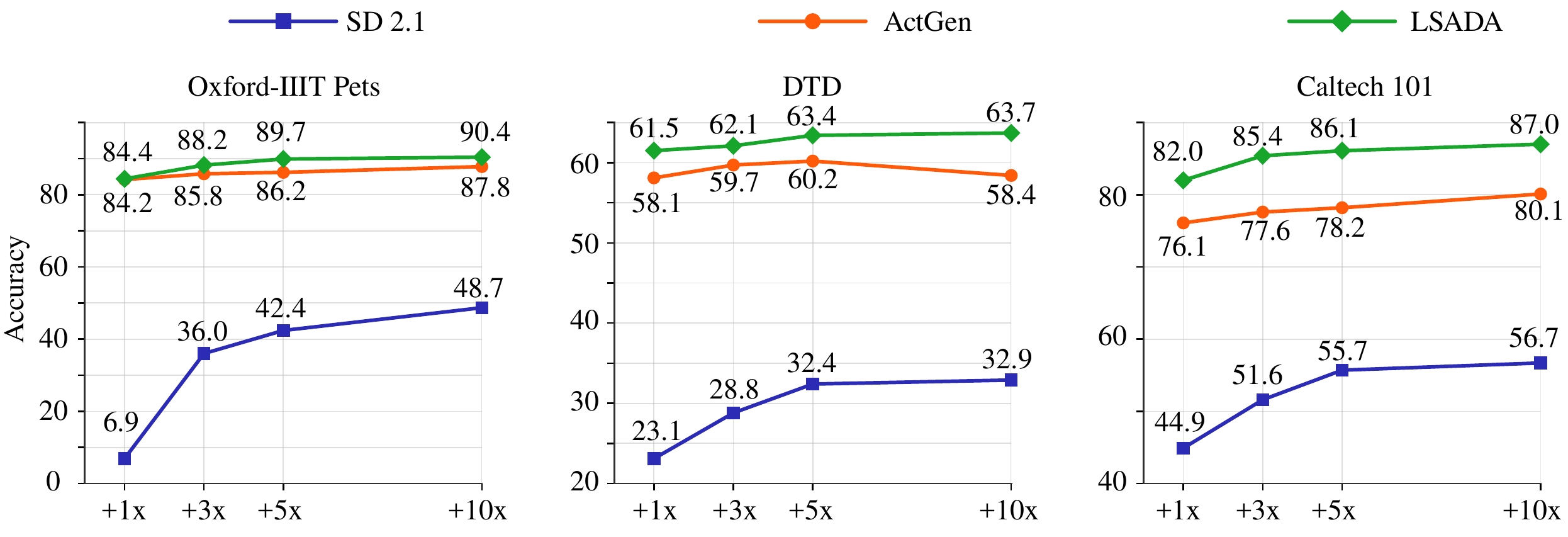}
   \caption{Accuracy (in \%) of a ResNet-50 trained from scratch on the original and generated datasets by LSADA, SD~2.1 and ActGen with different generation ratios.}
   \label{fig:efficiency_LSADA}
\end{figure*}

As shown in Table~\ref{tab:overall_performance}, our method consistently outperforms rule-based data augmentation methods and the SOTA static GDA method by a large margin.
It also achieves higher classification accuracy than existing dynamic GDA methods.
In particular, LSADA surpasses ActGen, the SOTA dynamic GDA method, by an average of 4.5\% on natural image datasets and 2.5\% on medical image datasets.
These results suggest that selecting samples solely based on model predictions may provide insufficient guidance in small-scale data scenarios.
In contrast, LSADA characterizes each sample using its learning state, maps it to a sample-specific augmentation strength, and employs a generation strategy adapted to that strength, making it well suited to small-scale data scenarios.

\subsubsection{DA efficiency.}
We further compare LSADA with ActGen and SD~2.1 under different generation ratios.
As shown in Fig.~\ref{fig:efficiency_LSADA}, LSADA consistently achieves competitive or higher accuracy using fewer generated images than ActGen.
In particular, on DTD and Caltech-101, LSADA with a $1\times$ generation ratio achieves higher accuracy than ActGen with a $10\times$ ratio.
Thus, LSADA requires only one-tenth as many generated images on these datasets, demonstrating at least a $10\times$ improvement in data augmentation efficiency.

\begin{table}[t]
    \centering
    {
    \begin{tabular}{cccc}
        \toprule
        Method & ResNeXt-50 & WideResNet-50 & MobileNet-V2 \\
        \midrule
        Original & 18.4 & 32.0 & 26.2 \\
        ActGen & 84.6 & 86.5 & 79.5 \\
        LSADA & \textbf{89.9} & \textbf{90.4} & \textbf{85.1} \\
        \bottomrule
    \end{tabular}
    }
    \caption{Accuracy of various architectures trained on 5×-generated Pets.}
    \label{tab:model_genera}
\end{table}

\subsubsection{Generalization to various architectures.}
In practical applications, generated images may be used to train downstream models with various architectures.
To evaluate the architectural generalizability of LSADA, we use the $5\times$-generated images obtained with feedback from ResNet-50 to train ResNeXt-50, WideResNet-50, and MobileNet-V2 from scratch.
As shown in Table~\ref{tab:model_genera}, LSADA consistently improves the classification accuracy of all evaluated architectures.
These results show that, although model feedback is used during data generation, the resulting images are not specific to the guiding architecture and can effectively transfer to other downstream models.

\begin{table}[!t]
    \centering
    {
    \begin{tabular}{cccc}
        \toprule
        Feedback & Current Loss & Loss-Decrease & Acc \\
        \midrule
        No Feedback & $\times$ & $\times$ & 88.9 \\ 
        Prediction Feedback & $\times$ & $\times$ & 89.2 \\
        Loss Only & \checkmark & $\times$ & 89.1 \\
        Decrease Only & $\times$ & \checkmark & 89.0 \\
        Learning State (Ours) & \checkmark & \checkmark & \textbf{89.7} \\
        \bottomrule
    \end{tabular}
    }
    \caption{Accuracy (in $\%$) comparison of different learning state signals on Pets.}
    \label{tab:ablation_ls}
\end{table}

\begin{table}[!t]
    \centering
    {
    \begin{tabular}{ccc}
        \toprule
        Mapping strategies &  Ours & Reversed \\
        \midrule
        Acc & \textbf{89.7} & 88.6 \\
        \bottomrule
    \end{tabular}
    }
    \caption{Accuracy (in $\%$) comparison under different mapping strategies on Pets.}
    \label{tab:ablation_map}
\end{table}

\begin{table}[!t]
    \centering
    \begin{tabular}{cccc}
        \toprule
        Generation strategies &  Ours & I2I & Paste \\
        \midrule
        Acc & \textbf{89.7} & 86.2 & 88.7 \\
        \bottomrule
    \end{tabular}
    \caption{Accuracy (in $\%$) comparison under different generation strategies on Pets.}
    \label{tab:ablation_generation}
\end{table}

\subsection{Ablation Study}
To further evaluate the effectiveness of LSADA, we conduct ablation studies on key components, including the learning state feedback, the mapping from learning state to augmentation strength, and the decoupled data augmentation and diffusion fusion generation strategy.

\subsubsection{Learning state feedback.}

To evaluate the effectiveness of the proposed feedback, we keep all other components fixed and vary only the feedback signal.
The compared variants include no feedback, the prediction-based feedback used in ActGen, current loss only, and loss-decrease rate only.
As shown in Table~\ref{tab:ablation_ls}, combining the current loss and loss-decrease rate achieves the highest accuracy, validating the effectiveness of the proposed learning state.

\subsubsection{Mapping strategy.}
To evaluate the effectiveness of mapping the learning state $L_i^t$ to the augmentation strength $a_i^t$, we compare the proposed strategy with a reversed mapping.
The proposed strategy assigns weaker augmentation to samples with larger learning state values and stronger augmentation to those with smaller values, while the reversed strategy follows the opposite assignment.
As shown in Table~\ref{tab:ablation_map}, the proposed mapping consistently achieves higher accuracy, demonstrating the effectiveness of adapting augmentation strength to the learning state of each sample.

\subsubsection{Decoupled data augmentation and diffusion fusion generation.}
To evaluate the effectiveness of the proposed generation strategy, we compare it with global image-to-image diffusion generation~(I2I) and decoupled augmentation followed by direct pasting~(Paste).
As shown in Table~\ref{tab:ablation_generation}, our strategy achieves the highest accuracy when all other components are kept unchanged.
These results demonstrate that decoupled data augmentation and diffusion fusion better balance image diversity and class-semantic preservation than global editing or direct pasting.

We further compare image quality across generation strategies.
As shown in Table~\ref{tab:quality}, I2I has a much higher FID, indicating a larger distribution gap from the original images.
Paste achieves the lowest FID, likely because it directly preserves the foreground.
Overall, the results suggest that an appropriate distribution gap may benefit downstream performance.

\begin{table}[t]
    \centering
    {
    \begin{tabular}{cccc}
        \toprule
        Generation strategies &  FID $\downarrow$ & IS $\uparrow$ & LPIPS $\downarrow$ \\
        \midrule
        I2I & 107.8 & 10.33 & 0.71 \\
        Paste & 40.10 & 17.58 & 0.74 \\
        Ours & 46.61 & 14.75 & 0.73 \\
        \bottomrule
    \end{tabular}
    }
    \caption{Image quality comparison under different generation strategies on Pets.}
    \label{tab:quality}
\end{table}

\begin{figure}[!t]
    \centering
    \includegraphics[width=.6\linewidth]{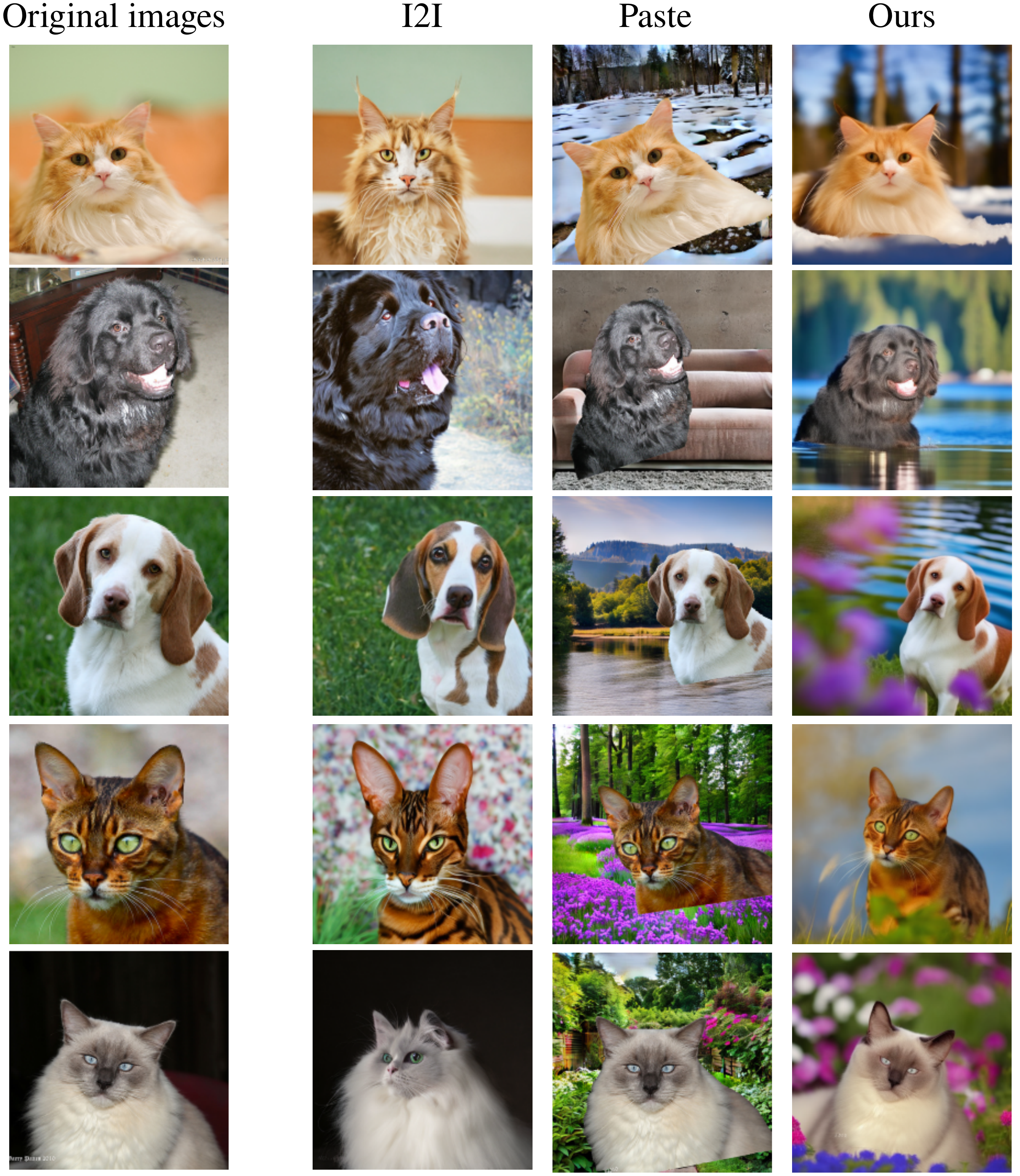}
    \caption{Natural image visualization under different generation strategies.}
    \label{fig:visual}
\end{figure}

The samples of natural datasets generated by the three strategies are visualized in Fig.~\ref{fig:visual}.
Although the I2I strategy produces diverse images, it often alters class-relevant regions, leading to semantic distortion.
The Paste strategy better preserves class semantics, but directly pasting the class-relevant region onto the background makes the images look unnatural, which may cause them to deviate from the data manifold.
In contrast, our strategy preserves semantic consistency while achieving more natural fusion between the class-relevant region and the background.
Similar observations on the medical image datasets are provided in Appendix~D.

\section{Conclusion}

This paper addresses two key limitations of existing dynamic GDA methods: the difficulty of determining sample-specific augmentation strengths and the lack of region-specific generation strategies for balancing image diversity and class semantics.
To address these limitations, we propose LSADA, which characterizes the learning state of each sample using its current loss and loss-decrease rate and adaptively determines its augmentation strength.
Furthermore, LSADA introduces a decoupled augmentation and diffusion fusion generation strategy to achieve region-specific augmentation, improving diversity while preserving class semantics.
Extensive experiments demonstrate the effectiveness of LSADA on both natural and medical image classification tasks.

There are several directions for future research.
For instance, extending LSADA to other computer vision tasks, such as object detection and semantic segmentation, would be valuable.
Additionally, exploring how learning-state guidance can be integrated with other generative models and data modalities is an interesting avenue for future investigation.



\bibliographystyle{unsrt}  
\bibliography{references}  

\newpage

\appendix
\section{Algorithm}

The complete algorithm is summarized in Algorithm~\ref{alg:lsada}.
LSADA alternates between downstream classifier optimization and learning-state-aware image generation.
At each epoch, the classifier is updated using the union of the original and previously generated data.
Every $k$ epochs, LSADA computes the current loss and loss-decrease rate of each original sample and combines them to estimate its learning state.
The learning states are then normalized and mapped to sample-specific augmentation strengths.
For each sample, LSADA extracts the class-relevant region and applies a strength-controlled rule-based transformation to both the region and its mask.
Meanwhile, we use an LLM to generate diverse background prompts, which guide the diffusion fusion process in Eqs.~(10)--(17) to progressively integrate the transformed class-relevant region with a generated class-irrelevant region~(background prompt).
The resulting images are added to $\mathcal{D}_g$ and used in subsequent classifier updates.
After $T$ augmentation epochs, the final classifier is trained on $\mathcal{D}_o\cup\mathcal{D}_g$.

\begin{algorithm}[H]
\caption{The LSADA Framework}
\label{alg:lsada}
\textbf{Input}: Original dataset $\mathcal{D}_o=\{(x_i,y_i)\}_{i=1}^{n}$; generative model $G$, classification model $f$; maximum augmentation epochs $T$; update interval $k$; generation number $m_t$ for each sample at epoch $t$, rule-based augmentation $\mathrm{T}$.

\textbf{Output}: Generated dataset $\mathcal{D}_g$ and final classification model $f_{\theta}$ parameterized by $\theta$.
\begin{algorithmic}[1]
\STATE $\mathcal{D}_g\leftarrow\emptyset$
\FOR{$t=1,\ldots,T$}
    \STATE Update the classification model to obtain $f_{\theta^t}$ using  $\mathcal{D}_o\cup\mathcal{D}_g$ 
    \IF{$t\bmod k=0$}
        \FOR{$i=1,\ldots,n$}
            \STATE Compute current loss: $\ell_i^t\leftarrow
                \mathcal{L}(f_{\theta^t}(x_i),y_i)$
            \STATE Compute loss-decrease rate: $\delta_i^t\leftarrow
                (\ell_i^{t-k}-\ell_i^t)/(\ell_i^{t-k}+\epsilon)$;
            \STATE Compute learning state value: $L_i^t\leftarrow\ell_i^t-\beta\delta_i^t$
        \ENDFOR
        \STATE Normalize $\{L_i^t\}_{i=1}^n$ to obtain
            $\{\widetilde{L}_i^t\}_{i=1}^n$
        \FOR{$i=1,\ldots,n$}
            \STATE Compute augmentation strength: $a_i^t\leftarrow a_{\min}+
                (a_{\max}-a_{\min})(1-\widetilde{L}_i^t)$
            \STATE Extract class-relevant region: mask $M_i\leftarrow\mathrm{Segment}(x_i)$;
                $x_i^r\leftarrow M_i\odot x_i$
                
            \STATE Rule-based Transform on class-relevant region: $(\hat{x}_i^r,\hat{M}_i)\leftarrow
                \mathrm{T}(x_i^r,M_i;a_i^t)$
            \STATE Generate prompts: $\{p_i^j\}_{j=1}^{m_t}\leftarrow
                \mathrm{LLM}(x_i,y_i,m_t)$
            \FOR{$j=1,\ldots,m_t$}
                \STATE Generate image using Eqs.~(10)--(17)\\ $\hat{x}_{i,j}^t\leftarrow
                    \mathrm{DiffusionFusion}
                    (G,\hat{x}_i^r,\hat{M}_i,p_i^j)$
                \STATE $\mathcal{D}_g\leftarrow\mathcal{D}_g
                    \cup\{(\hat{x}_{i,j}^t,y_i)\}$
            \ENDFOR
        \ENDFOR
    \ENDIF
\ENDFOR
\STATE $\theta\leftarrow
    \mathrm{TrainClassifier}(\mathcal{D}_o\cup\mathcal{D}_g)$
\STATE \textbf{return} $\mathcal{D}_g$ and $f_{\theta}$
\end{algorithmic}
\end{algorithm}

\section{Datasets}

We evaluate our method on nine publicly available image classification datasets, including six natural image datasets and three medical image datasets. A detailed summary of their statistics is provided in Table~\ref{tab:dataset_statistics}.

\begin{table*}[!t]
    \centering
    
    {
    \begin{tabular}{lcccc}
        \toprule
        \textbf{Dataset}
        & \textbf{Type}
        & \textbf{Classes}
        & \textbf{Samples}
        & \textbf{Average samples per class} \\
        \midrule
        Caltech 101       & Natural & 102 & 3,060  & 30    \\
        CIFAR100-Subset   & Natural & 100 & 10,000 & 100   \\
        Stanford Cars     & Natural & 196 & 8,144  & 42    \\
        Flowers           & Natural & 102 & 6,552  & 64    \\
        Pets              & Natural & 37  & 3,842  & 104   \\
        DTD               & Natural & 47  & 3,760  & 80    \\
        \midrule
        BreastMNIST       & Medical & 2   & 78     & 39    \\
        PathMNIST         & Medical & 9   & 10,004 & 1,112 \\
        OrganSMNIST       & Medical & 11  & 13,940 & 1,267 \\
        \bottomrule
    \end{tabular}
    }
    \caption{Statistics of both natural and medical image datasets.}
    \label{tab:dataset_statistics}
\end{table*}

The natural image datasets include Caltech 101~\cite{fei2004learning},
CIFAR100-Subset (CIFAR100-S)~\cite{krizhevsky2009learning},
Stanford Cars~\cite{krause2013collecting},
Oxford 102 Flowers (Flowers)~\cite{nilsback2008automated},
Oxford-IIIT Pets (Pets)~\cite{parkhi2012cats},
and DTD~\cite{cimpoi2014describing}.
CIFAR100-Subset is constructed from CIFAR100 by randomly sampling 100 images from each class, resulting in 10,000 images across 100 categories. This controlled subset provides a representative small-scale setting for evaluating the effectiveness of our method.
Collectively, these datasets cover multiple image recognition scenarios, including coarse-grained object classification (\textit{i.e.}, Caltech 101 and CIFAR100-Subset), fine-grained object
classification (\textit{i.e.}, Stanford Cars, Flowers, and Pets), and texture classification (\textit{i.e.}, DTD).

The medical image datasets are obtained from the MedMNIST
benchmark~\cite{yang2021medmnist}, including
BreastMNIST~\cite{al2020dataset},
PathMNIST~\cite{kather2019predicting},
and OrganSMNIST~\cite{xu2019efficient}.
These datasets contain breast ultrasound, colon pathology, and abdominal CT images, respectively.
To construct small-scale training scenarios, we use the official validation splits of BreastMNIST and PathMNIST as their training sets. 
For OrganSMNIST, we retain the original training split because its available training set is already relatively limited.

\section{Implementation Details}

\paragraph{Training and generation schedule.}
The downstream classifier is trained for 200 epochs, while image generation is performed only during the first $T=100$ augmentation training epochs.
With $k=5$, the number of generation rounds is $R=\lfloor T/k\rfloor=20$, and the generation epochs are
$\mathcal{T}_g=\{5,10,\ldots,100\}$ under one-based epoch
indexing.
We set the overall generation ratio to $m=5$, yielding a
target per-round generation ratio of $m/R=0.25$.

This fractional ratio is realized at the dataset level. 
At each generation epoch, approximately $n/4$ original samples are randomly selected without replacement from $\mathcal{D}_o$, and one image is generated for each selected sample.
The 20 generation rounds therefore yield approximately $5n$ generated images in total, although no individual original sample is required to be selected exactly five times.
For each selected sample, its learning state determines the augmentation strength $a_i^t$.
The generated samples are accumulated in $\mathcal{D}_g$ and used together with $\mathcal{D}_o$ in subsequent training epochs.
After epoch 100, no additional images are generated, and training continues on $\mathcal{D}_o\cup\mathcal{D}_g$ until epoch 200.

\paragraph{Computing environment.}
All experiments are conducted using four NVIDIA RTX 4090 GPUs,
each with 24 GB of memory, with PyTorch 2.1.2 and CUDA 12.1.

\paragraph{Diffusion configuration.}
Let $N$ denote the total number of DDIM inference steps. 
We set $N=10$, $\tau_1=\lfloor0.5N\rfloor=5$, and
$\tau_2=\lfloor0.7N\rfloor=7$.
In each cycle, noise is injected from $\tau_1$ to $\tau_2$, followed by reverse denoising with masked injection from $\tau_2$ to $\tau_1$.
Under the implementation's inference-step indexing, masked injection is applied at timesteps $\{7,6,5\}$.
This cycle is repeated 10 times.

During the final reverse denoising stage from $\tau_1$ to $0$, we
additionally perform masked injection at a dataset-specific set of
late-stage timesteps, denoted by $\mathcal{T}_{\mathrm{inj}}$. We
set $\mathcal{T}_{\mathrm{inj}}=\{4,3\}$ for Caltech-101, Cars,
Flowers, Pets, and DTD, and
$\mathcal{T}_{\mathrm{inj}}=\{4\}$ for CIFAR100-Subset,
PathMNIST, BreastMNIST, and OrganSMNIST.
At each $\tau\in\mathcal{T}_{\mathrm{inj}}$, the corresponding inverted class-relevant latent is injected into the class-relevant region.
Apart from these additional timesteps, the remaining reverse
denoising steps follow Eq.~(16) in the main paper without masked
injection.

\section{Visualization on Medical Datasets}

\begin{figure}[H]
    \centering
    \includegraphics[
        width=.6\columnwidth,
        trim=4.9cm 0.6cm 5.2cm 0.8cm,
        clip
    ]{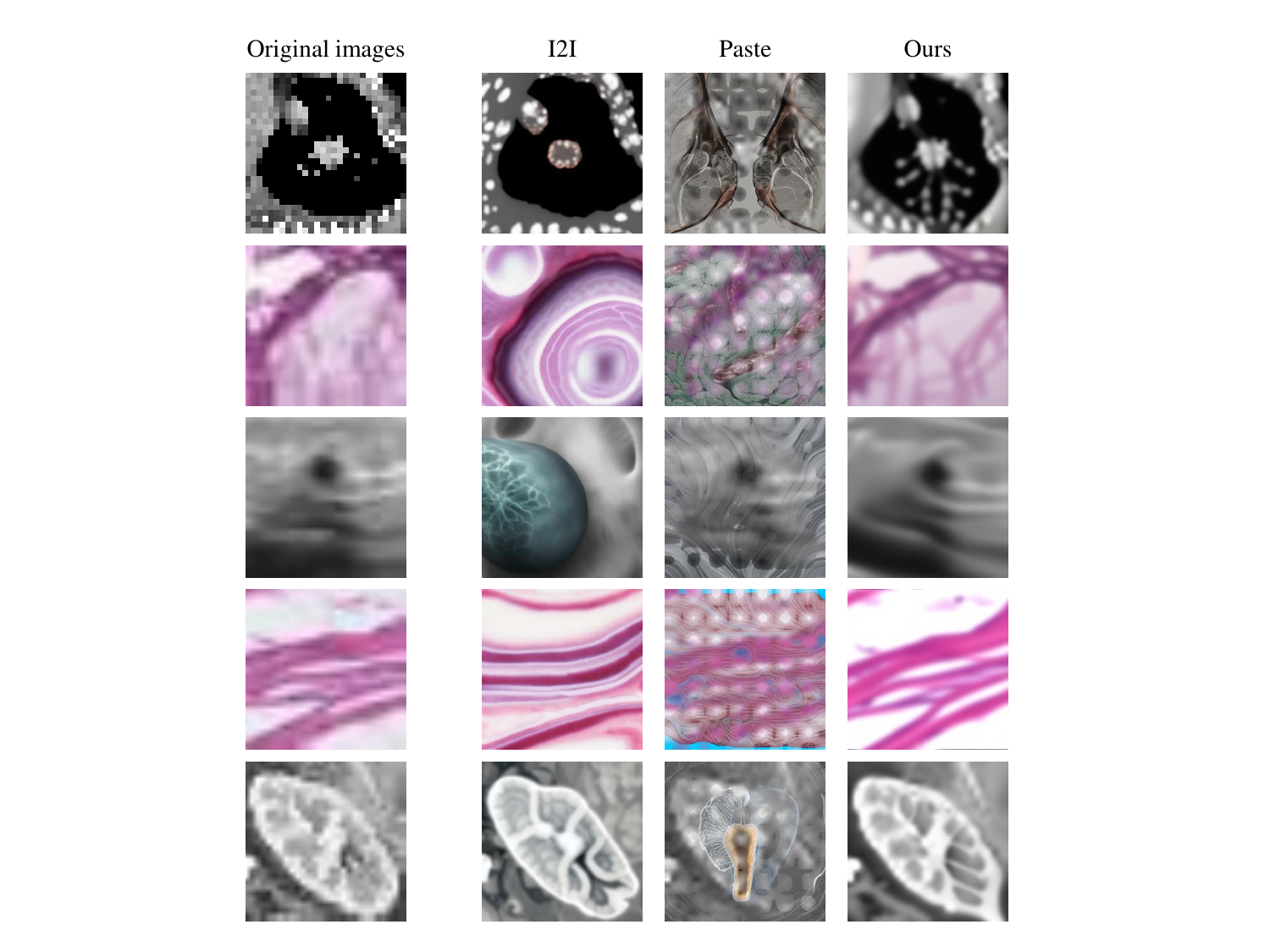}
    \caption{Medical image visualization under different generation strategies.}
    \label{fig:medical_visualization}
\end{figure}

The samples generated by the three strategies on the medical image datasets are visualized in Fig.~\ref{fig:medical_visualization}.
Compared with the original images, I2I generally preserves the main medical structures but introduces noticeable changes in global appearance and texture, which may cause semantic distortion.
Paste and our method both use the same CAM maps to preserve class-related regions.
However, Paste directly blends the preserved regions with generated backgrounds in pixel space, which may lead to inconsistent structures and textures and produce unnatural visual patterns.
In contrast, our diffusion fusion process integrates the preserved and generated regions more smoothly, producing visually coherent images while retaining discriminative medical structures.
Overall, our method achieves a better balance between semantic preservation and visual quality.

\end{document}